\documentclass{article}

\usepackage[preprint]{neurips_2025}

\usepackage[utf8]{inputenc} 
\usepackage[T1]{fontenc}    
\usepackage{hyperref}       
\usepackage{url}            
\usepackage{booktabs}       
\usepackage{amsfonts}       
\usepackage{nicefrac}       
\usepackage{microtype}      
\usepackage{xcolor}         
\usepackage{tabularx}
\usepackage{makecell}
\usepackage{multirow}
\usepackage{graphicx}
\usepackage{adjustbox}
\usepackage{subcaption}
\usepackage{caption}
\usepackage{wrapfig}
\usepackage{float}
\usepackage{amsmath}
\usepackage[table]{xcolor}
\usepackage{sidecap, caption}
\newcolumntype{H}{>{\setbox0=\hbox\bgroup}c<{\egroup}@{}}

\title{SNCE: Geometry-Aware Supervision for Scalable Discrete Image Generation}

\author{Shufan Li$^{1,2,*}$, Jiuxiang Gu$^{1}$, Kangning Liu$^{1}$, Zhe Lin$^{1}$, \textbf{Aditya Grover$^{2}$, Jason Kuen$^{1}$} \\
$^1$Adobe~~$^2$UCLA \\
* Work done primarily during internship at Adobe Research \\
}

\begin{document}

\maketitle

\begin{abstract}
 Recent advancements in discrete image generation showed that scaling the VQ codebook size significantly improves reconstruction fidelity. However, training generative models with a large VQ codebook remains challenging, typically requiring larger model size and a longer training schedule. In this work, we propose Stochastic Neighbor Cross Entropy Minimization (SNCE), a novel training objective designed to address the optimization challenges of large-codebook discrete image generators. Instead of supervising the model with a hard one-hot target, SNCE constructs a soft categorical distribution over a set of neighboring tokens. The probability assigned to each token is  proportional to the proximity between its code embedding and the ground-truth image embedding, encouraging the model to capture semantically meaningful geometric structure in the quantized embedding space. We conduct extensive experiments across class-conditional ImageNet-256 generation, large-scale text-to-image synthesis, and image editing tasks. Results show that SNCE significantly improves convergence speed and overall generation quality compared to standard cross-entropy objectives.

\end{abstract}



\section{Introduction}
\label{sec:intro}

Modern image generation models achieve high-fidelity synthesis by first encoding raw image pixels into low-dimensional latent embeddings \cite{podell2023sdxl,esser2024scaling-sd3,xie2025sana,xie2025sana1,flux2024}. Compared with directly modeling image pixels, training models to generate these latent embeddings has proven to be significantly more effective and scalable. The most widely adopted approach for latent image generation is the latent diffusion model (LDM), which trains a neural network to generate latent image embeddings from i.i.d.\ Gaussian noise through a continuous diffusion process \cite{rombach2022high}. 

Recently, discrete image generation models \cite{hu2022unified,bai2024meissonic,chang2022maskgit} have drawn increasing attention due to their compatibility with discrete language-modeling architectures, making them attractive candidates for unified multimodal models \cite{yang2025mmada,li2025lavidao}. In addition, these models demonstrate efficiency advantages because they support key–value (KV) caching during generation \cite{li2025sparse,ma2025dkv}.

Unlike LDMs, which directly learn to generate continuous latent representations, discrete image generation models first discretize continuous image latents into discrete tokens and then train a neural network to generate sequences of such tokens. In general, discrete image generation models can be categorized into two families: autoregressive (AR) models and discrete diffusion models. AR models generate image tokens sequentially in a left-to-right order, whereas discrete diffusion models begin with a sequence consisting entirely of special mask tokens and gradually unmask them to recover clean image tokens.
\begin{figure}[t]
    \centering
    \includegraphics[width=1.0\linewidth]{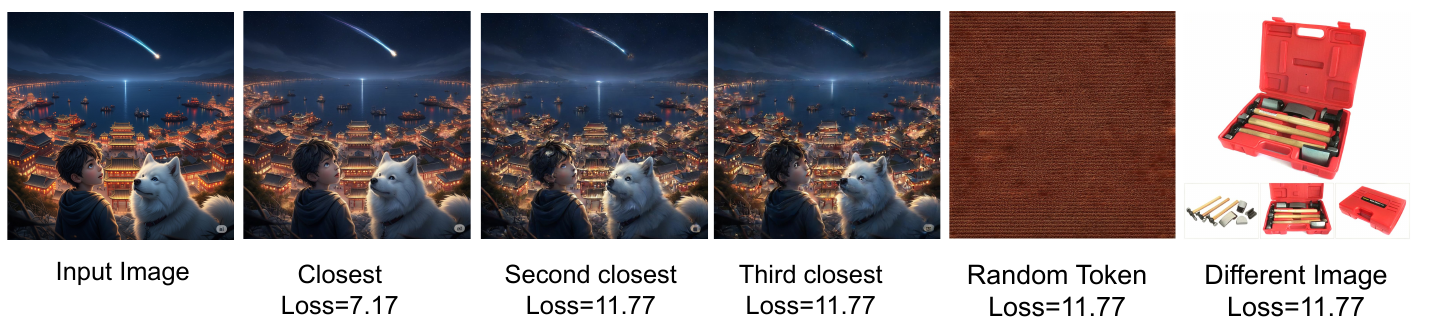}
    \caption{\textbf{Limitations of Vanilla Cross Entropy Loss with One-hot Target.} Given an input image, vanilla CE loss cannot distinguish between non-closet tokens in the embedding space, even though some of the tokens are close to ground truth in embedding space and can decode to semantically similar images. Addtional details of loss computation in thig figure can be found in appendix.}
    \label{fig:visual_distance}
\end{figure}

Most discrete image generators, including both AR models and discrete diffusion models, share two common design choices. First, they rely on a discrete image tokenizer that quantizes continuous image latents into discrete tokens. This is typically implemented using a vector quantization (VQ) module. Second, they employ cross-entropy (CE) loss as the training objective to learn a categorical distribution over the vocabulary of possible tokens, although the exact formulation differs slightly between AR models and discrete diffusion models.



The quality of the image tokenizer plays a critical role in determining the fidelity of generated images. Several works have shown that larger vocabulary size (codebook size) leads to better reconstruction quality since a larger vocabulary is more expressive and can better capture fine-grained details in the image \cite{shi2025scalable,zhu2024scaling}. However, training image generators with a large codebook size can be difficult since it requires larger model size and more data than training image generators with a small codebook.

We refer to this challenge as the \textbf{codebook sparsity problem}. As the vocabulary size grows, the frequency of each token during training decreases substantially, resulting in increasingly sparse supervision signals for individual tokens. This can be understood by considering token frequencies in the training data. Consider a dataset of $1$M images where each image is represented by $256$ tokens. Assuming uniform token frequencies, each token in the codebook will appear on average $31{,}250$ times for an $8{,}192$-sized codebook, but only $1{,}280$ times for a $200$K-sized codebook. Hence, the per-token learning signal becomes much sparser as the vocabulary grows, making optimization more difficult.


Although this sparsity may appear analogous to language modeling, where the vocabulary size is similarly large, the challenge is fundamentally different because image modeling is inherently a high-entropy problem. For example, when training unified models that generate both images and text \cite{cui2025emu3}, the cross-entropy loss for image tokens ($>7.0$) is significantly larger than that for language tokens ($<1.0$). Intuitively, given an English sentence with only the final few words missing, the probability mass typically concentrates on a small set of plausible candidates. In contrast, even when only a small region of an image is missing (e.g., the eye region of a portrait), there exist many plausible pixel configurations that could complete the image. As a result, the prediction distribution in image modeling is inherently more diffuse, making learning with sparse supervision substantially more challenging. 

A major contributing factor to this issue is that the standard cross-entropy loss uses one-hot probability vectors as training targets, assigning all probability mass to a single ground-truth token while treating all other tokens equally as incorrect. We argue that this formulation is unnatural for tokens produced by a VQ tokenizer. In the VQ process, the quantizer maintains a code embedding for each token in the vocabulary. During tokenization, a continuous image latent embedding is compared with all code embeddings, and the closest one is selected as the ground-truth token according to a similarity metric such as L2 distance or cosine similarity. However, the cross-entropy loss does not distinguish among non-ground-truth tokens: the second-best candidate, the third-best candidate, and a completely unrelated token are all treated identically. This limitation is illustrated in Figure~\ref{fig:visual_distance}. This issue becomes particularly severe in  large-codebook tokenizers, where two highly similar image patches can map to different tokens, making one-hot supervision increasingly brittle.



To address the optimization challenges of discrete image generation with large vocabularies, we propose \textbf{stochastic neighborhood cross-entropy (SNCE)} minimization. The key insight of this work is that supervision for discrete image tokens should respect the geometry of the underlying VQ embedding space rather than treating tokens as independent categorical labels. Instead of using one-hot targets corresponding to the nearest codebook entry, SNCE constructs a soft categorical distribution over the vocabulary based on the distances between code embeddings and the encoded image latent. Tokens whose embeddings are closer to the latent representation receive higher probability in the target distribution. This design alleviates the codebook sparsity problem by allowing multiple nearby tokens in the embedding space to receive positive learning signals, rather than supervising the model using only a single closest token.

To validate the effectiveness of SNCE, we conduct small-scale experiments on ImageNet-256 and large-scale experiments on text-to-image generation and image editing. Our results show that SNCE significantly improves both convergence speed and final generation fidelity compared with standard CE training. Overall, these findings suggest that incorporating embedding-space geometry into the training objective is crucial for scaling discrete image generators to large vocabularies, and that SNCE serves as a promising drop-in replacement for vanilla CE in discrete image generation models with large codebooks.

\section{Background and Related Works}

\subsection{Discrete Image Tokenizer}

Discrete image tokenizers encode images into sequences of discrete codes. 
VQ-VAE~\cite{van2017neural} first introduced a vector quantization (VQ) module that converts continuous features into discrete tokens via a learnable codebook. 
Several subsequent works improved image fidelity through multiscale hierarchical architectures~\cite{razavi2019generating}, adversarial training objectives~\cite{esser2021taming}, and residual quantization~\cite{lee2022autoregressive}. 
However, naively scaling the codebook size and latent dimension of these models often leads to low code utilization and latent collapse. 
VQGAN-LC~\cite{zhu2024scaling} mitigates collapse by using a frozen codebook. 
FSQ~\cite{mentzer2023finite} improves utilization by reducing the latent dimension. 
LFQ~\cite{yu2023language} set the codebook embedding dimension to zero and scales the codebook size to $2^{18}$ through factorization. 
However, these approaches do not fundamentally resolve the quantization bottleneck when the latent dimension is large. 
IBQ~\cite{shi2025scalable} first achieves high utilization for large codebooks with high-dimensional latents through index backpropagation. 
FVQ~\cite{shi2025scalable} further introduces a VQ-bridge module to simultaneously scale both the codebook size and the latent dimension.

Formally, a canonical image tokenizer consists of a continuous image encoder $F_{\text{enc}}$ and a vector codebook $V$. 
The encoder $F_{\text{enc}}$ maps image pixels $x \in \mathbb{R}^{H \times W \times 3}$ to continuous latents 
$z \in \mathbb{R}^{L \times D}$, where $L$ denotes the number of latent tokens and $D$ is the latent dimension. 
The codebook $V = \{v_i\}_{i=1}^{K}$ is a set of vectors $v_i \in \mathbb{R}^{D}$, where $K$ denotes the codebook size. 
During quantization, each latent vector $z_i$ is compared with all code vectors using a distance metric $d(\cdot)$, and the index of the closest code is selected as the discrete representation. 
This process can be written as

\begin{align}
z &= F_{\text{enc}}(x), \nonumber \\
y_i &= {\text{argmin}}_{k \in \{1,\dots,K\}} d(z_i, v_k), 
\quad \forall i \in \{1,\dots,L\}.
\label{eq:vq_enc}
\end{align}

While recent advances in image tokenizers have significantly improved the scalability of the codebook, training a discrete image generator with a large codebook remains challenging due to the optimization issues discussed in Section~\ref{sec:intro}. 
This work focuses on addressing this bottleneck by introducing a carefully designed training objective SNCE.

\subsection{Discrete Image Generation}

Discrete image generators learns to generate discrete image tokens produced by a tokenizer as opposed to directly modeling continuous latents or raw pixels. They can be categorized into two classes: autoregressive models and discrete diffusion models.

\textbf{Autoregressive models} generate $L$ tokens in a left-to-right sequential order. VQVAE\cite{van2017neural} and VQGAN \cite{esser2021taming} first explored autoregressive image generation. DALL-E\cite{} scaled autoregressive models to large-scale text-to-image generation via a prior model.  Parti \cite{yu2022scaling} scaled the model size to 20B parameters for high-fidelity generation.  Llama-Gen \cite{sun2024autoregressive} draw inspiration from language models and applied the Llama \cite{touvron2023llama} architecture to image generation tasks. Most recently, several works such as Janus \cite{chen2025janus} and Emu-3 \cite{wang2024emu3} explored training unified autoregressive model for both visual understanding and generation tasks, demonstrating that discrete image generation can achieve comparable performance to state-of-the-art continuous diffusion models and is a promising approach for building unified multi-modal models.

Autoregressive image generators employ the same next-token-prediction objective as their counterparts in the language domain during training. Given a sequence $y=[y_1,..y_L]$ with $L$ discrete tokens, an autoregressive model $p_{\theta}$ is trained by minimizing the following objective

\begin{align}
    \mathcal{L}_{\text{AR}}=\mathbb{E}_y[-\sum_{i=1}^L\log p_\theta(y_i|y_1..y_{i-1})]
    \label{eq:ar}
\end{align}
\textbf{Discrete diffusion models} generate multiple tokens in parallel at each step, making them more efficient than autoregressive models~\cite{lou2023discrete-sedd,sahoo2024simple}. 
Given a sequence of image tokens $y^0 = [y_1^0, \dots, y_L^0]$, the forward discrete diffusion process $q(y^t|y^s)$ gradually converts clean tokens in $y^0$ into a special mask token $[\text{M}]$ over the continuous time interval $[0,1]$, where $0 \le s \le t \le 1$. 
A neural network $\Theta$ parameterizes the reverse process $p_\theta(y^s|y^t)$. 
At inference time, we initialize the sequence $y^1 = [\text{M}, \dots, \text{M}]$ as a fully masked sequence. 
We then gradually unmask these tokens over the time interval $[0,1]$ by repeatedly invoking the learned reverse process $p_\theta(y^s|y^t)$ over multiple diffusion steps until we obtain a sequence of clean tokens $y^0$. 

MaskGIT~\cite{chang2022maskgit} first explored this form of masked image generation. 
Meissonic~\cite{bai2024meissonic} incorporated several architectural innovations, such as token compression, and scaled generation to $1024 \times 1024$ resolution. 
More recent works have explored building unified understanding and generation models using the discrete diffusion paradigm, including MMaDa~\cite{yang2025mmada}, the LaViDa-O series~\cite{li2025lavidao,li2025sparse,li2026lavida}, and Unidisc~\cite{hu2022unified}. 
These works demonstrate that discrete diffusion models is a more promising approach for large-scale visual generation tasks than AR models.

During training, given a clean sequence $y^0 = [y_1, \dots, y_L]$, a partially masked sequence $y^t$ is sampled from the forward diffusion process $q(y^t|y^0)$. 
We then optimize the model prediction $p_\theta(y^0|y^t)$ by minimizing the negative ELBO:

\begin{equation}
\mathcal{L}_{\text{ELBO}} =
\mathbb{E}_{y^0,\, t \sim \text{Unif}([0,1]),\, y^t \sim q(y^t|y^0)}
\left[
-\frac{1}{t} \sum_{i=1}^{L} 
\mathbf{I}\{y_i^t = [\text{M}]\}
\log p_\theta(y_i^0 \mid y^t)
\right]
\label{eq:mdm}
\end{equation}

where $\mathbf{I}\{y_i^t = [\text{M}]\}$ is a binary indicator function that equals 1 if $y_i^t = [\text{M}]$.

While there have been considerable advances in large-scale training of discrete image generators, most existing works remain constrained by the optimization challenges associated with large codebooks and therefore adopt tokenizers with relatively small codebooks. 
The only exception is Emu3.5~\cite{cui2025emu3}, which trains a large discrete diffusion model with a codebook size of 131{,}072 by scaling the model to 30B parameters and leveraging massive training data.

We argue that the main bottleneck for scaling the codebook size lies in the common likelihood term $\log p_\theta$, which appears in both autoregressive and discrete diffusion objectives. 
Since this term is implemented as the cross-entropy loss between predicted per-token logits and a one-hot target vector, it leads to weak per-token training signals when the codebook size becomes large. 
In this work, we explore how to leverage the superior image fidelity of large-codebook tokenizers while overcoming this optimization challenge with the proposed SNCE objective.

\subsection{Soft Labels}

Using soft targets in cross-entropy loss has been widely studied in the context of classification problems. 
Label smoothing mixes uniform vectors with one-hot targets~\cite{lukasik2020does} to regularize training and mitigate label noise. 
It has been widely applied in many areas, including image classification~\cite{muller2019does}, image segmentation~\cite{islam2021spatially}, and graph learning~\cite{zhou2023adaptive}.  Knowledge distillation~\cite{zhou2021rethinking} is another common form of soft labeling, where a student network is trained using soft labels generated by a teacher network. 
Several works have explored using soft labels derived from the agreement and confidence scores of human annotators~\cite{wu2023don,singh2025soft}. 
Other works treat soft labels as learnable parameters and optimize them through meta-learning~\cite{vyas2020learning}. 

Applying soft labeling to discrete image generation remains relatively underexplored, beyond a few works on model distillation~\cite{zhu2025di}. 
Our work is the first to design a soft-label training objective for discrete image generation that explicitly addresses the token sparsity issue caused by large codebook sizes.
\section{Method}

\subsection{Stochastic Neighbor Embedding}

The concept of stochastic neighbors was first introduced as a component of t-distributed Stochastic Neighbor Embedding (t-SNE) \cite{van2008visualizing}, which visualizes high-dimensional vectors in a 2D space while preserving their high-dimensional structure. Given $N$ high-dimensional vectors $v_1,\dots,v_N$, it defines the pairwise neighborhood distribution $p_{j|i}$ for each $(i,j)\in\{1,2,\dots,N\}^2$ as

\begin{align}
p_{j|i} =
\begin{cases}
\dfrac{\exp\!\left(-\lVert x_i-x_j\rVert^2 / 2\sigma_i^2\right)}
{\sum_{k\ne i}\exp\!\left(-\lVert x_i-x_k\rVert^2 / 2\sigma_i^2\right)}, & j \ne i, \\[6pt]
0, & j=i .
\end{cases}
\end{align}

where the bandwidth $\sigma_i$ is chosen via binary search to match a target perplexity. This ensures that each point considers a similar number of neighbors, which improves the quality of the resulting visualization.

Inspired by this formulation, we design a categorical neighbor distribution for image latents. Recall from Equation~\ref{eq:vq_enc} that a discrete image tokenizer first encodes an image $x \in \mathbb{R}^{H\times W \times 3}$ into continuous latents $z \in \mathbb{R}^{L \times D}$. The tokenizer then quantizes $z$ using a codebook $V=\{v_1,\dots,v_K\}$. For each continuous latent $z_i$, we define a neighborhood distribution over the $K$ tokens as

\begin{align}
q_k(z_i) =\dfrac{\exp\!\left(-d(z_{i},v_k) / 2\tau^2\right)}
{\sum_{j=1}^K\exp\!\left(-d(z_{i},v_j) / 2\tau^2\right)}, 
\quad \forall k \in\{1,2,\dots,K\}
\label{eq:sne_img}
\end{align}

where $\tau$ is a fixed hyperparameter and $d(\cdot)$ is the distance metric used by the tokenizer during vector quantization. Common choices for $d(\cdot)$ include the L2 distance, negative cosine similarity, and negative dot product. In our experiments, we adopt the IBQ tokenizer, which uses the negative dot product as the dissimilarity metric (i.e., $d(x,y)=-x^\top y$). We also exprimented with FVQ tokenizer, with uses L2 distance (i.e., $d(x,y)=\lVert x- y\rVert^2$). We set $\tau=0.71$ in our setup. Additional ablation studies on the choice of hyperparameters are provided in the appendix.

There are two key differences compared with vanilla t-SNE. First, in the standard t-SNE formulation, pairwise neighborhood probabilities are defined over a finite set of vectors, and the probability of a vector being its own neighbor is set to zero. In our setup, we instead compute neighborhood probabilities between an arbitrary continuous vector $z_i$ and a finite set of codebook vectors $v_1,\dots,v_K$. When $z_i = v_r$ for some $r \in \{1,\dots,K\}$ (which occurs for synthetic images produced by discrete tokenizers or pre-tokenized images), the probability $q_r(z_i)=q_r(v_r)$ is not zero. Instead, it attains the highest value among all $K$ indices, which is desirable since the closest code should receive the highest probability in the training targets.

Second, t-SNE uses a per-sample bandwidth $\sigma_i$ determined via binary search. For efficiency reasons, this procedure is impractical during training. We therefore replace it with a fixed temperature shared across all points. This choice also better aligns with the nature of learnable codebooks, whose density may vary across the embedding space. If a latent vector is close to many code vectors, those indices should naturally receive higher probabilities; conversely, if it is close to only a few codes, we should not artificially increase the temperature to enforce a fixed number of neighbors.

\subsection{Stochastic Neighbor Cross Entropy Loss}

Both the autoregressive objective in Equation~\ref{eq:ar} and the discrete diffusion objective in Equation~\ref{eq:mdm} share a common term $\log p_\theta (y_i|\cdot)$, which denotes the model's predicted log-probability of the ground-truth token $y_i$. This term is typically implemented as the negative cross-entropy loss in the following form

\begin{align}
   J_\text{CE}= \log p_\theta (y_i|\cdot) = \sum_{k=1}^K \textbf{I}\{y_i=k\} \log p_\theta (Y_i=k|\cdot)
\end{align}

where $Y_i$ is the random variable corresponding to the ground-truth token $y_i$, and $\log p_\theta (Y_i=k|\cdot)$ denotes the predicted log-probability. The indicator $\textbf{I}\{y_i=k\}$ is a one-hot target.

In our proposed SNCE objective, we replace the one-hot vector with the neighborhood distribution $q_k(z_i)$ defined in Equation~\ref{eq:sne_img}, leading to the following objective

\begin{align}
   J_\text{SNCE} &= \sum_{k=1}^K  q_k (z_i) \log p_\theta (Y_i=k|\cdot)  \\
   & =\sum_{k=1}^K \dfrac{\exp\!\left(-d(z_{i},v_k) / 2\tau^2\right)}
{\sum_{j=1}^K\exp\!\left(-d(z_{i},v_j) / 2\tau^2\right)} \log p_\theta (Y_i=k|\cdot)
\end{align}

For both autoregressive models and discrete diffusion models, we can use $J_\text{SNCE}$ as a drop-in replacement for $J_\text{CE}$. The only difference lies in the conditional term inside $\log p_\theta (Y_i=k|\cdot)$. In autoregressive models, the term $\log p_\theta (Y_i=k|y_1,\dots,y_{i-1})$ is conditioned on a prefix sequence, whereas in discrete diffusion models the term $\log p_\theta (Y_i^0=k|y^t)$ is conditioned on a partially masked sequence. We offer three interpretations for this modification.

\textbf{Categorical Variational Autoencoder.} Continuous VAEs encode images into a distribution (typically a diagonal Gaussian) rather than a deterministic embedding. In contrast, VQ-VAE and its variants are deterministic and produce a fixed sequence of codes for each image. The SNCE objective can be interpreted as modifying the quantization process so that each token $y_i$ is not determined by selecting the nearest codebook vector, but instead is sampled from the categorical neighbor distribution defined in Equation~\ref{eq:sne_img}. Taking the autoregressive model as an example, we obtain

\begin{align}
    \mathbb{E}_{z\sim\mathcal{D}}&[-\sum_{i=1}^L J_{\text{SNCE}}(q(z_i),\theta )] = 
       \mathbb{E}_{z\sim\mathcal{D},y_i\sim q(z_i)}[-\sum_{i=1}^L J_{\text{CE}}(y_i, \theta )]
    \label{eq:ar_eq1}
\end{align}

Several works, such as RobustTok \cite{qiu2025robust}, demonstrate that stochastic quantization (e.g., sampling from the top-$k$ closest tokens) can make generative model training more robust and improve generation quality. Compared with such explicit stochastic quantization methods, SNCE is equivalent in expectation but has lower variance and is more stable because it directly operates on the probability vector $q(z_i)$ rather than on Monte Carlo samples $y_i\sim q(z_i)$. Moreover, explicit sampling does not address the low token-frequency issue unless many candidates are sampled for each $q(z_i)$.

\textbf{Knowledge Distillation with KL Divergence Minimization.} Knowledge distillation is typically used to transfer knowledge from a larger teacher model to a smaller student model. However, several works such as Reverse Distillation \cite{nasser2024reverse} and Weak-to-Strong Generalization \cite{ildiz2024high,burns2312weak} show that a weaker teacher can also improve the training of a stronger model by accelerating convergence and improving generalization. The proposed SNCE loss can be viewed as minimizing the KL divergence between a weak teacher model (the discrete tokenizer) and a strong student model (the generator). Concretely, the neighborhood distribution $q(z_i)$ can be viewed as a teacher that encodes a continuity inductive bias: tokens that are close in the latent space should have similar probabilities. This is a reasonable assumption, as Figure~\ref{fig:visual_distance} shows that token embedding distance correlates with the semantic distance between decoded images.

Since KL divergence can be decomposed into a cross-entropy term and an entropy term, we have

\begin{align}
    \min_\theta \mathbb{E}_z[\mathbb{D}_{KL}(q(z_i)\lVert p_\theta(Y_i|\cdot))] 
    &= \min_\theta \mathbb{E}_z[H(q(z_i), p_\theta(Y_i|\cdot)) - H(q(z_i))] \nonumber \\
     &= \min_\theta\mathbb{E}_z[-J_{\text{SNCE}}(q(z_i),\theta)]
     \label{eq:kl_div}
\end{align}

where $H(q(z_i))$ is the entropy term independent of $p_\theta$ and can therefore be ignored during optimization.

\textbf{On-Policy Learning with Continuous Rewards.} Another line of work formulates token prediction as a decision-making problem where a policy $\pi(a|s)$ selects an action $a$ given a state $s$. The action space corresponds to the vocabulary, and each action corresponds to a token. The state $s$ may represent a prefix sequence in autoregressive generation or a partially masked sequence in a discrete diffusion process. Wu et al.~\cite{wu2025diversity} show that the gradient of the cross-entropy objective is equivalent to policy gradients with reward 
\[
r_\text{CE}(s, a)=\frac{\textbf{I}\{a=y_i\}}{p_\theta(Y_i=a|\cdot)} .
\]
The numerator encourages exploitation of the ground-truth signal, while the denominator encourages exploration by penalizing tokens that already have high probability.

When replacing CE with SNCE, the reward becomes
\[
r_\text{SNCE}(s, a)=\frac{q_a(z_i)}{p_\theta(Y_i=a|\cdot)},
\]
where the binary indicator $\textbf{I}\{a=y_i\}$ is replaced by the smooth value $q_a(z_i)$ from Equation~\ref{eq:sne_img}, which depends on embedding distance. Figure~\ref{fig:visual_distance} shows that embedding distance is a strong surrogate for image reconstruction quality. Therefore, incorporating it into the reward provides a more informative training signal than a binary indicator.

\subsection{Gradient Analysis}

In practice, the model predicts unnormalized logits $h=[h_1,\dots,h_K]\in\mathbb{R}^K$ through a final linear projection layer at each token position. The log probabilities are obtained via the log-softmax operator
\[
\log p_\theta(Y_i=k\mid\cdot)=\text{logSoftmax}(h)_k .
\]

Given a target probability vector $w\in\mathbb{R}^K$, the gradient of the objective with respect to the logits $h$ is

\begin{align}
    \frac{\partial J}{\partial h_k}=w_k - p_\theta(Y_i=k\mid\cdot),
    \label{eq:gradient}
\end{align}

where $p_\theta(Y_i=k \mid \cdot)\in(0,1)$ denotes the predicted probability of token $k$.

Under the standard cross-entropy (CE) objective, the supervision weight is $w_k=\mathbf{I}\{y_i=k\}$. Thus, only the ground-truth token receives a positive update, while every other token $k\neq y_i$ receives a negative gradient $\frac{\partial J}{\partial h_k}=-p_\theta(Y_i=k\mid\cdot).$



This penalty becomes stronger when the model assigns large probability to non-ground-truth tokens. In large-codebook discrete image models, tokens that are close in the embedding space often correspond to visually similar reconstructions. By continuity, a well-trained model may assign relatively high probability to tokens near the closest token. However, the CE objective strongly penalizes these semantically similar alternatives even when the reconstructed image remains visually faithful (Figure~\ref{fig:visual_distance}). This creates unnecessary optimization difficulty and further exacerbates token-frequency imbalance, since rare but semantically similar tokens rarely receive positive updates.

In contrast, SNCE replaces the one-hot supervision with the soft distribution $q(z_i)$. Instead of concentrating all supervision on a single token, it distributes learning signals across neighboring tokens in the embedding space. As a result, semantically similar tokens receive positive gradients proportional to their proximity to $z_i$, which leads to smoother optimization dynamics and mitigates the token-frequency imbalance by allowing more tokens to receive positive updates during training.

\section{Experiments}

\subsection{Validation on a Toy Example}

\begin{figure}[t]
    \centering
    \includegraphics[width=1.0\linewidth]{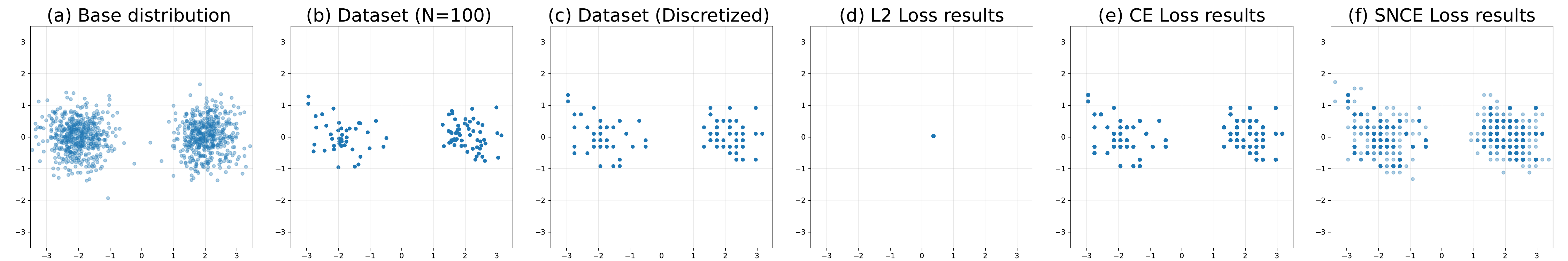}
    \caption{\textbf{Toy examples on 2D Gaussians.} (a) ground truth distribution (b) 100-sample dataset (c) discretized dataset (d) L2 regression results (e) CE loss results (f) SNCE loss results}
    \label{fig:toy}
\end{figure}

We first validate our design using a simple toy example. We assume the image latent consists of a single continuous embedding $(L=1)$ in a 2D space $(D=2)$. The ground-truth latent distribution is defined as a mixture of two Gaussians, shown in Figure~\ref{fig:toy}(a). We use a simple quantization scheme consisting of a $50\times50$ uniform grid, where the quantization process maps a continuous latent to the closest grid point. We sample 100 points from the ground-truth distribution as our dataset, shown in Figure~\ref{fig:toy}(b). This setup is analogous to a large vocabulary with limited training data. Figure~\ref{fig:toy}(c) visualizes the quantized dataset.

Since there is only one token, there is no distinction between autoregressive and discrete diffusion models in this example. We therefore train a 10-layer MLP with a constant input and evaluate three objectives: (1) continuous L2 regression, (2) cross-entropy (CE), and (3) SNCE. The model is trained until convergence, and the learned distributions are visualized in Figure~\ref{fig:toy}(d,e,f).

Setup (1) with L2 loss collapses to a single point near the mean of the two ground-truth Gaussians. Setup (2) with CE loss perfectly fits the training data but fails to capture the underlying distribution due to the codebook sparsity problem arising from a large vocabulary and limited data. In contrast, Setup (3) with SNCE better approximates the ground-truth distribution by assigning non-zero probability to neighboring tokens. This occurs because SNCE provides positive training signals for nearby tokens rather than only the closest one.

We draw two insights from these results. First, although SNCE resembles a discrete surrogate for a regression objective, directly replacing the cross-entropy loss with a regression loss is ineffective when the ground-truth distribution is multi-modal in the embedding space. Second, compared with standard CE with one-hot targets, SNCE injects a continuity inductive bias into training, enabling the model to better approximate the ground-truth distribution in large-vocabulary settings with limited data.

\subsection{Validation on ImageNet256}

Our second validation experiment is conducted on ImageNet \cite{russakovsky2015imagenet} $256\times256$ class-conditioned image generation. We adopt the Emu3.5 tokenizer with a codebook size of 131,072, as well as FVQ \cite{chang2025scalable} with a codebook size of 262,144. The generator follows the architecture of IBQ-B, an autoregressive transformer with 342M parameters. The only architectural modifications are the input embedding layer and the final linear projection head, whose sizes are increased to match the larger codebook. We train the model for 100 and 300 epochs using both objectives and report FID scores over 50k sampled images. During sampling, we sweep classifier-free guidance (CFG) values from 2.5 to 5.5 and report the best result for each run. The results are summarized in Table~\ref{tab:snce_results}.

The results show that SNCE outperforms CE by both accelerating convergence and achieving better image fidelity measured by FID. An interesting observation is that simply increasing the codebook size from 16,384 to 131,072 introduces an additional 235M parameters to the model, corresponding to a 68\% increase in total parameters. Moreover, these parameters—primarily in the final linear projection layer—receive mostly negative training signals when trained with the standard CE objective. Although SNCE consistently outperforms CE, the final FID score remains higher than that of the baseline model with the smaller codebook. This highlights the optimization challenges of training generative models with extremely large vocabularies under limited data. Nevertheless, SNCE provides significant improvements over the vanilla CE baseline.

\begin{table}[t]
\centering
\caption{\textbf{Class-conditioned Image Synthesis on ImageNet256 datset.}  *Model have identical-sized transformer layer. Parameter count increased due to larger token embedding and final linear head.}
 \label{tab:snce_results}
\scriptsize
\begin{tabular}{lcccccc}
\toprule
\textbf{Model} & \textbf{Params} & \textbf{Tokenizer} &\textbf{Tokenizer Pretraining} &\textbf{Codebook} & \textbf{Epoch} & \textbf{FID} $\downarrow$ \\
\midrule
IBQ-B & 342M  & IBQ \cite{shi2025scalable} & ImageNet256 & 16,384 & 300 & 2.88 \\
\midrule
\rowcolor{gray!20}
IBQ-B-Ours (CE)   & 577M* &  Emu3.5-IBQ \cite{cui2025emu3} & Large-Scale T2I & 131,072 & 100 & 7.53 \\
\rowcolor{gray!20}
IBQ-B-Ours (SNCE) & 577M* & Emu3.5-IBQ \cite{cui2025emu3}  & Large-Scale T2I  &131,072 & 100 & 3.62 \\
\midrule
\rowcolor{gray!20}
IBQ-B-Ours (CE) &  577M*  & Emu3.5-IBQ \cite{cui2025emu3} & Large-Scale T2I &131,072 & 300 & 5.44 \\
\rowcolor{gray!20}
IBQ-B-Ours (SNCE) & 577M* & Emu3.5-IBQ \cite{cui2025emu3} & Large-Scale T2I & 131,072 & 300 & 3.42\\
\midrule
\rowcolor{gray!20}
IBQ-B-Ours (CE) & 846M* & FVQ \cite{zhu2024scaling} & ImageNet256 & 262,144 & 300 & 4.11 \\
\rowcolor{gray!20}
IBQ-B-Ours (SNCE) & 846M*& FVQ \cite{zhu2024scaling} &ImageNet256 & 262,144 & 300 & 
3.20 \\
\bottomrule
\end{tabular}
\end{table}

\subsection{Text-to-Image Generation}

We then scale training to $1024\times1024$ high-resolution text-to-image synthesis on a dataset containing 50M images. We adopt a transfer learning framework and initialize from LaViDa-O, a 10B discrete diffusion model with unified multimodal understanding and generation capabilities. The original LaViDa-O tokenizer uses a codebook of size 8,192. We replace it with the Emu3.5 tokenizer \cite{cui2025emu3}, which contains 131,072 codes. The input embedding layer and the final linear projection layer are reinitialized to match the new vocabulary size, while all other parameters remain unchanged. The model is then fine-tuned for 200k steps. Additional training details are provided in the appendix.

We report text-to-image performance on the GenEval benchmark \cite{ghosh2023geneval} in Table~\ref{tab:geneval} and the DPG benchmark \cite{hu2024equipdpg} in Table~\ref{tab:dpg}. GenEval evaluates high-level text-to-image alignment, while DPG evaluates fine-grained alignment using dense captions that contain detailed descriptions. To evaluate image fidelity, we also report FID scores on MJHQ-30k \cite{li2024playground}. Since FID relies on a network trained on relatively low-resolution images and does not fully capture high-resolution image fidelity, we additionally report the HPSv3 score \cite{ma2025hpsv3}. HPSv3 is a vision-language reward model designed for evaluating high-resolution text-to-image generation and is aligned with human preferences. It has been shown to be effective at assessing perceptual image quality. These results are also summarized in Table~\ref{tab:dpg}.

Across all evaluation metrics, SNCE demonstrates superior performance compared with the standard CE objective, highlighting the effectiveness of soft targets at scale. Most notably, SNCE leads to substantial improvements in FID ($-3.67$) and HPSv3 ($+0.12$).

      
        

\begin{table*}[t]
\centering
\caption{\textbf{Text to Image Generation Performance on GenEval Dataset}. Cont. refers to continuous latent diffusion models.}
\label{tab:geneval}
\resizebox{1.0\linewidth}{!}{
\setlength{\tabcolsep}{3pt} 
{\begin{tabular}{lcccccccccHHH}
\toprule
 & \textbf{Parms} & \textbf{Codebook} & \textbf{Single}$\uparrow$  & \textbf{Two} $\uparrow$ & \textbf{Position}$\uparrow$  & \textbf{Counting}$\uparrow$  & \textbf{Color}$\uparrow$  & \textbf{Attribution}$\uparrow$  & \textbf{GenEval}$\uparrow$  & \textbf{Latency}$\downarrow$  & \textbf{Speedup}$\uparrow$ & \textbf{DPG}$\uparrow$ \\
 \midrule
         SDXL\cite{podell2023sdxl} & 3B  & Cont. & 0.98 & 0.74 & 0.39 & 0.85 & 0.15 & 0.23 & 0.55  & 5.2 & - & 84.0\\
        DALLE 3\cite{openai_dalle3} & - & Cont. &0.96 &  0.87 & 0.47 & 0.83 & 0.43 & 0.45 &  0.67 & -  &83.5 & 84.1 \\
        SD3\cite{esser2024scaling-sd3}  & 8B & Cont.  & 0.99   & 0.94 & 0.72 & 0.89 & 0.33 & 0.60 & 0.74 &23.3 & -   & 83.5 \\
        Flux-Dev\cite{flux2024} & 12B & Cont.  & 0.99 & 0.85 & 0.74 & 0.79 & 0.21 & 0.48  & 0.68 & 31.6 & -  & -\\
        Playground v3\cite{li2024playground} &-  & Cont.& 0.99 & 0.95 & 0.72 & 0.82 & 0.50 & 0.54 & 0.76 & - & - & -\\
                BAGEL \cite{deng2025emerging}  & 14B  & Cont.& 0.99& 0.94 &0.64& 0.81 &0.88 &0.63 &0.82 & 45.1 & -  & -\\
                Show-o \cite{xie2024show} & 1B & 8,192  &   0.98 & 0.80 & 0.31 & 0.66 &  0.84 & 0.50 & 0.68 & * & -  & -\\
        MMaDa\cite{yang2025mmada}  & 8B  & 8,192& 0.99 & 0.76 & 0.20 & 0.61 & 0.84 & 0.37 & 0.63  & * & -   & 53.4\\

\midrule
LaViDa-O \cite{li2025lavidao}  & 10B  & 8,192 & 0.99  & 0.85 &  0.65 & 0.71 & 0.86 & 0.58 & 0.77 & 21.27 & 1.00 $\times$ & 81.8\\
\rowcolor{gray!20}

+Adaptation (CE) &10B  &131,072  & 0.99 & 0.93 & 0.47 & 0.68 & 0.88 & 0.55 & 0.74 & 10.86 & 1.95 $\times$ & 82.4 \\
\rowcolor{gray!20}
+Adapt (SNCE) &10B  & 131,072  & 1.00 & 0.95 & 0.63 & 0.68  & 0.87 & 0.57 & 0.78 & 10.86 & 1.95 $\times$  & 83.3\\
\bottomrule
\end{tabular}}}
\end{table*}

\begin{table}[t]
\centering
\caption{\textbf{Text-to-Image Generation Performance on DPG Benchmark and MJHQ-30k Dataset.}}
\label{tab:dpg}

\scriptsize
\setlength{\tabcolsep}{11pt}
\begin{tabular}{lcccrc}
\toprule
\multirow{2}{*}{\textbf{Model}} &
\multirow{2}{*}{\textbf{Params}} &
\multirow{2}{*}{\textbf{Codebook}} &
\multirow{2}{*}{\textbf{DPG}$\uparrow$} &
\multicolumn{2}{c}{\textbf{MJHQ-30k}} \\
\cmidrule(lr){5-6}
 &  &  &  & \textbf{FID}$\downarrow$ & \textbf{HPSv3}$\uparrow$ \\

\midrule
SD3\cite{esser2024scaling-sd3} & 8B & Cont. & 83.5 & 11.92 & - \\
Flux-Dev\cite{flux2024} & 12B & Cont. & - & 10.15 & -\\
Show-o \cite{xie2024show} & 1B & 8,192 & - & 15.18 & - \\
MMaDa\cite{yang2025mmada} & 8B & 8,192 & 53.4 & 32.85 & - \\
\midrule
LaViDa-O \cite{li2025lavidao} & 10B & 8,192 & 81.8 & 6.68 & 8.81 \\
\rowcolor{gray!20}
+Adaptation (CE) & 10B & 131,072 & 82.4 & 10.10 & 8.97\\
\rowcolor{gray!20}
+Adapt (SNCE) & 10B & 131,072 & 83.3 & 6.43 &  9.10 \\
\bottomrule
\end{tabular}
\end{table}

\subsection{Image Editing}

During the adaptation of LaViDa-O, we incorporate 2M image editing samples during the final 100k training steps to enable image editing capabilities. Notably, the scale of the editing dataset is significantly smaller than the text-to-image dataset, which may further exacerbate the low token-frequency issue associated with large codebooks. Additional training details are provided in the appendix.

We evaluate image editing performance on the ImgEdit benchmark \cite{ye2025imgedit}, which uses a GPT-4o \cite{openai2024gpt4o} judge model to produce evaluation scores. The results are reported in Table~\ref{tab:image-edit}. Both CE-based adaptation and SNCE-based adaptation improve performance over the small-codebook baseline. This improvement can largely be attributed to the higher reconstruction fidelity of the large-codebook tokenizer, which enables the model to better preserve fine-grained details from the input image. When directly comparing SNCE with CE, SNCE achieves a noticeable improvement in overall editing quality (+0.13).

\subsection{Additional Results}

\begin{figure}[t]
\centering
\begin{subfigure}{0.47\linewidth}
    \centering
    \includegraphics[width=\linewidth]{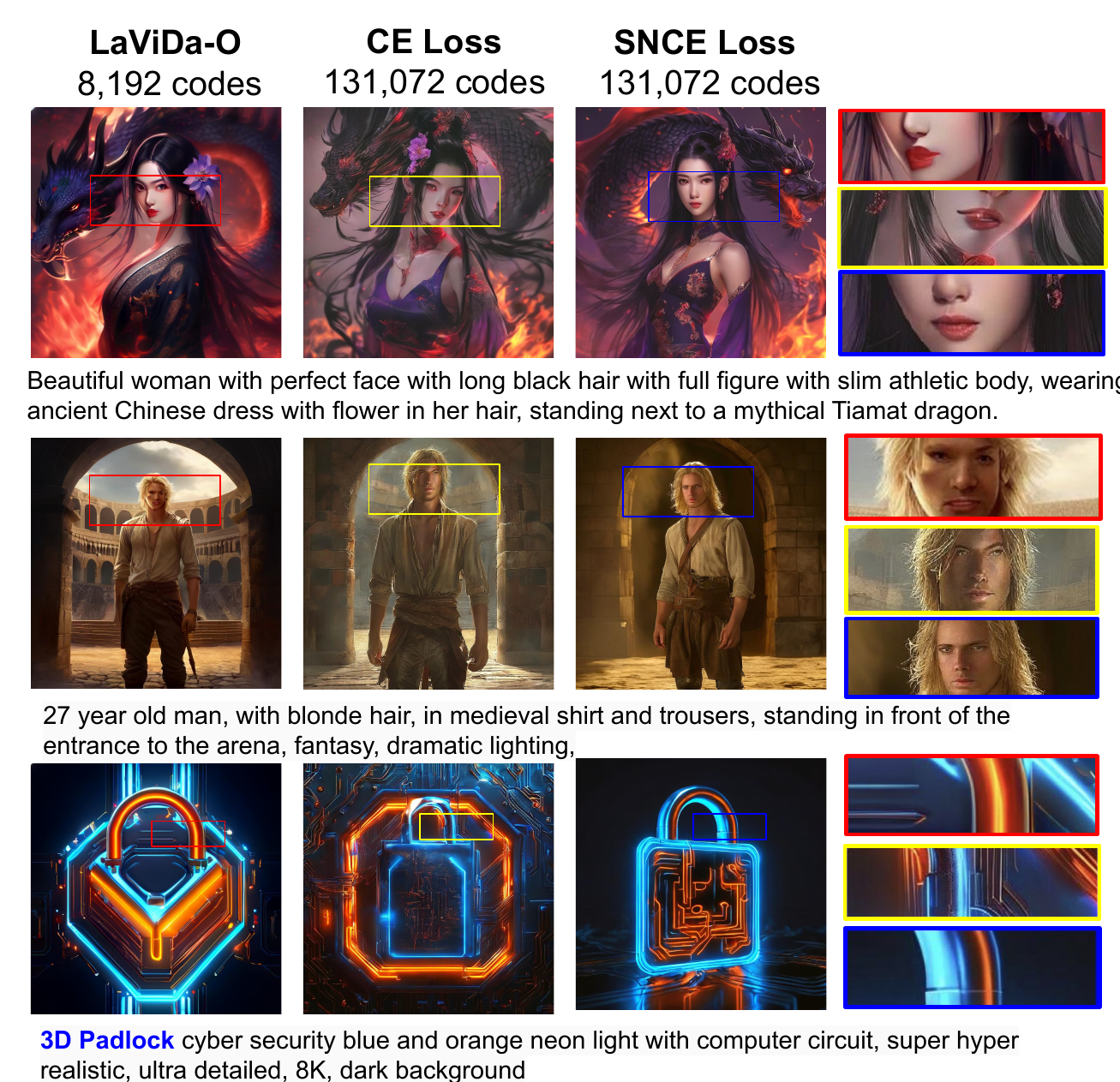}
    \caption{Text-to-Image Generation}
    \label{fig:demo_t2i}
\end{subfigure}
\hfill
\begin{subfigure}{0.49\linewidth}
    \centering
    \includegraphics[width=\linewidth]{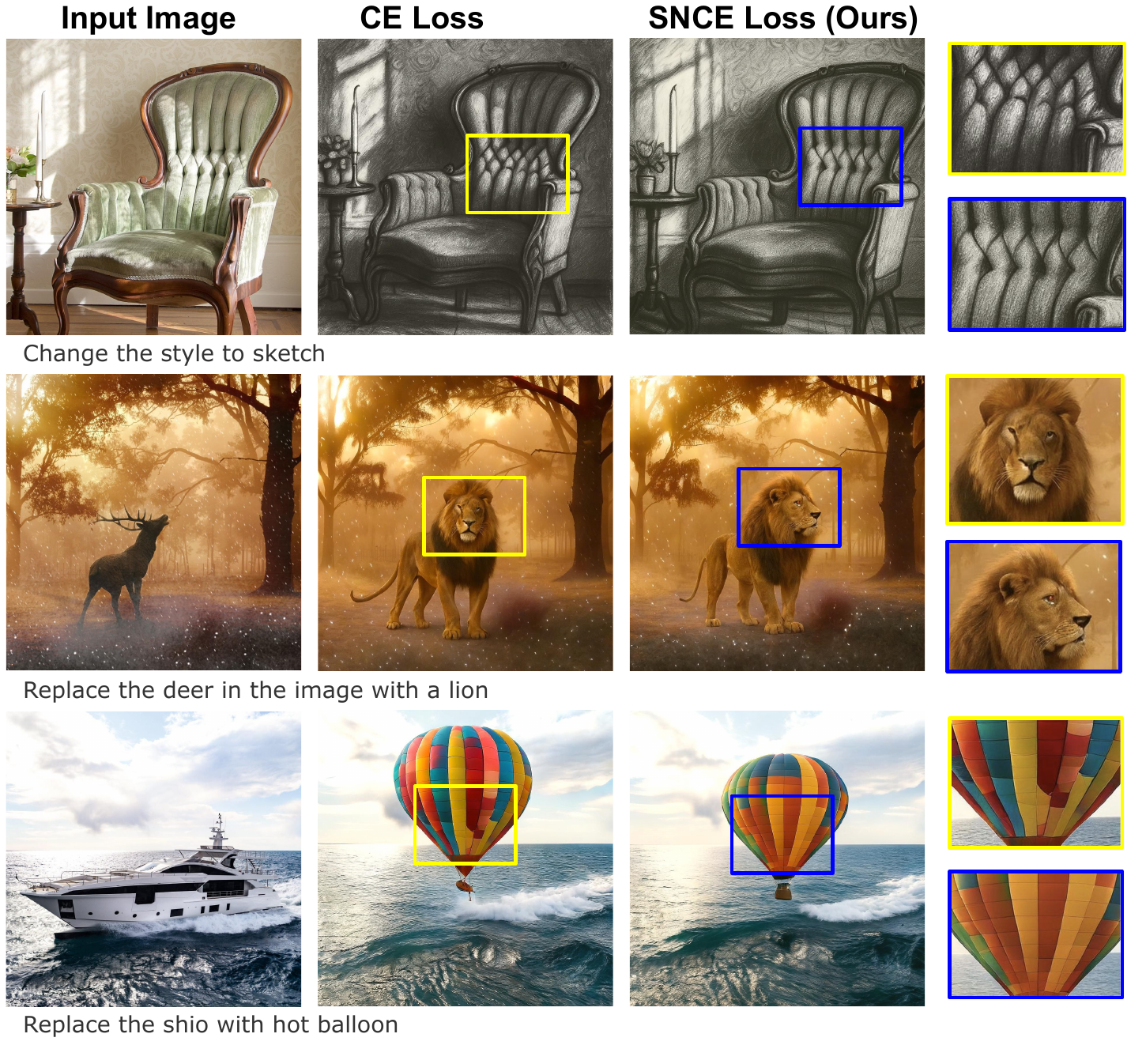}
    \caption{Image Editing}
    \label{fig:demo_edit}
\end{subfigure}
\caption{\textbf{Qualitative results.} (Left) Text-to-image generation. (Right) Image editing.}
\label{fig:demo}
\end{figure}

\textbf{Qualitative Comparison:} We present qualitative results for text-to-image generation in Figure~\ref{fig:demo_t2i} and image editing in Figure~\ref{fig:demo_edit}. For text-to-image generation, using a larger-codebook tokenizer produces images with more refined visual details. Furthermore, SNCE consistently outperforms CE in terms of text alignment, spatial structure, and the fidelity of low-level details. For image editing tasks, models trained with SNCE better preserve the input image structure while producing outputs with fewer artifacts. The appendix includes more examples. 

\textbf{Ablation Studies:} In the section of the main paper focus on main results and defer detailed ablation studies on hyperparameters to appendix.
\begin{table*}[h!]
\centering
\caption{\textbf{Image Editing Performance on ImgEdit benchmark.} }
\label{tab:image-edit}
\resizebox{1.0\linewidth}{!}{
\setlength{\tabcolsep}{3pt} 
{
\begin{tabular}{lccccccccccHH}
\hline
\textbf{Model} & \textbf{Add} $\uparrow$& \textbf{Adjust} $\uparrow$& \textbf{Extract}$\uparrow$ & \textbf{Replace}$\uparrow$ & \textbf{Remove}$\uparrow$ & \textbf{Background}$\uparrow$ & \textbf{Style}$\uparrow$ & \textbf{Hybrid}$\uparrow$ & \textbf{Action} $\uparrow$& \textbf{Overall}$\uparrow$  & \textbf{Latency}$\downarrow$ & \textbf{Speedup}$\uparrow$ \\
\hline
GPT-4o \cite{openai2024gpt4o} & 4.61 & 4.33 & 2.90 & 4.35 & 3.66 & 4.57 & 4.93 & 3.96 & 4.89 & 4.20 & 111.4 & -\\
Qwen2.5VL+Flux \cite{wang2025gpt} & 4.07 & 3.79 & 2.04 & 4.13 & 3.89 & 3.90 & 4.84 & 3.04 & 4.52 & 3.80 & 55.2 & -\\


FluxKontext dev \cite{labs2025flux1kontextflowmatching} & 3.76 & 3.45 & 2.15 & 3.98 & 2.94 & 3.78 & 4.38 & 2.96 & 4.26 & 3.52& 51.4 & - \\
OmniGen2 \cite{wu2025omnigen2}& 3.57 & 3.06 & 1.77 & 3.74 & 3.20 & 3.57 & 4.81 & 2.52 & 4.68 & 3.44 & 84.8 & -\\
UniWorld-V1 \cite{lin2025uniworld} & 3.82 & 3.64 & 2.27 & 3.47 & 3.24 & 2.99 & 4.21 & 2.96 & 2.74 & 3.26 & 56.2 & -\\
BAGEL \cite{deng2025emerging}& 3.56 & 3.31 & 1.70 & 3.30 & 2.62 & 3.24 & 4.49 & 2.38 & 4.17 & 3.20 & 88.2 & - \\
Step1X-Edit \cite{liu2025step1x}  & 3.88 & 3.14 & 1.76 & 3.40 & 2.41 & 3.16 & 4.63 & 2.64 & 2.52 & 3.06& - & - \\
OmniGen \cite{xiao2025omnigen1} & 3.47 & 3.04 & 1.71 & 2.94 & 2.43 & 3.21 & 4.19 & 2.24 & 3.38 & 2.96 & 126.2 & -\\
UltraEdit \cite{zhao2024ultraedit} & 3.44 & 2.81 & 2.13 & 2.96 & 1.45 & 2.83 & 3.76 & 1.91 & 2.98 & 2.70 & - & -\\
InstructAny2Pix\cite{li2023instructany2pix} & 2.55 & 1.83 & 2.10 & 2.54 & 1.17 & 2.01 & 3.51 & 1.42 & 1.98 & 2.12 & 48.2 & -\\ 
MagicBrush \cite{zhang2023magicbrush} & 2.84 & 1.58 & 1.51 & 1.97 & 1.58 & 1.75 & 2.38 & 1.62 & 1.22 & 1.90& - & - \\
Instruct-Pix2Pix\cite{brooks2023instructpix2pix} & 2.45 & 1.83 & 1.44 & 2.01 & 1.50 & 1.44 & 3.55 & 1.20 & 1.46 & 1.88 & 9.5 & -\\
\midrule
LaViDa-O \cite{li2025lavidao} & 4.04	&3.62	&2.01&	4.39	&3.98	&4.06	&4.82&	2.94 &	3.54&	3.71 & 63.98 & 1.00$\times$\\
\rowcolor{gray!20}
+Adaptation (CE) & 4.00 & 3.63 & 2.05 & 4.43 & 4.00 & 4.08 & 4.82 & 2.96 & 3.84 & 3.76    & 22.55 & 2.83$\times$ \\
\rowcolor{gray!20}
+Adaptation (SNCE) & 4.07	& 3.85&	2.46	& 4.53&	 4.00	&4.17 &		4.93	&2.90 &	4.08 & 3.89 & 22.55 & 2.83$\times$ \\

\hline
\end{tabular}
}
}
\end{table*}

\section{Conclusion}

We proposed SNCE, a simple yet effective modification to the training objective of discrete image generators. SNCE specifically addresses optimization bottlenecks in large-codebook settings by allowing positive training signals for tokens that are close, but not necessarily the closest, to the ground-truth latents in the embedding space.  We provided a detailed analysis of the advantages of SNCE, drawing connections to robust tokenization, weak-to-strong distillation, and on-policy learning. We further conducted extensive experiments to evaluate the effectiveness of SNCE, including small-scale validation studies as well as large-scale training for text-to-image generation and instruction-based image editing. Results across multiple benchmarks demonstrate that SNCE accelerates convergence and achieves improved image fidelity when training discrete image generators at scale compared with the standard CE objective. We hope that SNCE can facilitate the scaling of codebook sizes for building next-generation discrete image foundation models.

\bibliographystyle{plain}
\bibliography{main}
\clearpage

{\centering
\textbf{SNCE: Geometry-Aware Supervision for Scalable
Discrete Image Generation}\\
\vspace{0.5em}Appendix \\
\vspace{1.0em}}

\section{Additional Technical Details}

\subsection{Loss Implementation for AR and Discrete Diffusion Models}

In the main paper, we note that our method replaces the standard log-likelihood term $J_{\text{CE}}$ with the modified objective $J_{\text{SNCE}}$ in both autoregressive and discrete diffusion training objectives. In this section, we provide the concrete formulations of these two objectives.

\textbf{Autoregressive Loss.} Recall the next-token prediction objective for autoregressive models from Equation~\ref{eq:ar}:

\begin{align}
\mathcal{L}_{\text{AR}}=
\mathbb{E}_{y}\left[-\sum_{i=1}^L
\log p_\theta(y_i|y_1,\dots,y_{i-1})
\right]
\end{align}

We replace $\log p_\theta(y_i|y_1,\dots,y_{i-1})$ with $J_{\text{SNCE}}$, which yields

\begin{align}
\mathcal{L}_{\text{AR-SNCE}}
&=
\mathbb{E}_{z}\left[
-\sum_{i=1}^L
\sum_{k=1}^K
q_k(z_i)
\log p_\theta(Y_i=k|y_1,\dots,y_{i-1})
\right]
\nonumber\\
&=
\mathbb{E}_{z}\left[
-\sum_{i=1}^L
\sum_{k=1}^K
\frac{\exp\!\left(-d(z_i,v_k)/(2\tau^2)\right)}
{\sum_{j=1}^K \exp\!\left(-d(z_i,v_j)/(2\tau^2)\right)}
\log p_\theta(Y_i=k|y_1,\dots,y_{i-1})
\right]
\end{align}

\textbf{Discrete Diffusion Loss.} Recall the masked diffusion model (MDM) loss from Equation~\ref{eq:mdm}:

\begin{equation}
\mathcal{L}_{\text{ELBO}} =
\mathbb{E}_{y^0,\, t \sim \text{Unif}([0,1]),\, y^t \sim q(y^t|y^0)}
\left[
-\frac{1}{t}
\sum_{i=1}^{L}
\mathbf{I}_M
\log p_\theta(y_i^0 \mid y^t)
\right]
\end{equation}

where $\mathbf{I}_M = \mathbf{I}\{y_i^t = [\text{M}]\}$ indicates whether the token is masked.

Replacing $\log p_\theta(y_i^0|y^t)$ with $J_{\text{SNCE}}$ gives

\begin{align}
\mathcal{L}_{\text{ELBO-SNCE}}
=
\mathbb{E}_{z,\, t,\, y^t}
\Bigg[
-\frac{1}{t}
\sum_{i=1}^{L}
\mathbf{I}_M
\sum_{k=1}^K
q_k(z_i)
\log p_\theta(Y_i=k \mid y^t)
\Bigg]
\nonumber\\
=
\mathbb{E}_{z,\, t,\, y^t}
\Bigg[
-\frac{1}{t}
\sum_{i=1}^{L}
\mathbf{I}_M
\sum_{k=1}^K
\frac{\exp\!\left(-d(z_i,v_k)/(2\tau^2)\right)}
{\sum_{j=1}^K \exp\!\left(-d(z_i,v_j)/(2\tau^2)\right)}
\log p_\theta(Y_i=k \mid y^t)
\Bigg]
\end{align}

In both cases, we replace the expectation over quantized latents $y$ with an expectation over continuous latents $z$. Note that $y$ in the AR loss and $y^0$ in the discrete diffusion loss are deterministically obtained from $z$ via quantization, and therefore do not need to be explicitly included in the expectation.

\subsection{Categorical VAE Perspective}

In this section, we provide a more detailed derivation of the categorical VAE interpretation of the SNCE loss.

Given latent features $z_i$, the standard quantization process $Q(z_i)$ converts them into discrete tokens $y_i$ through

\begin{align}
y_i = Q(z_i) =
\text{argmin}_{k \in \{1,\dots,K\}}
d(z_i, v_k)
\end{align}

where $d$ denotes a distance metric. Several works, such as RobustTok~\cite{qiu2025robust}, have shown that introducing stochasticity into the quantization process can be beneficial. Suppose we instead use a tokenizer that independently samples tokens from the categorical distribution $q$ defined in Equation~\ref{eq:sne_img}:

\begin{align}
y_i \sim q(z_i)
\end{align}

Then we have

\begin{align}
\mathbb{E}_{z_i,\, y_i \sim q(z_i)}
[\log p(z_i|\cdot)]
=
\sum_{k=1}^K
q_k(z_i)
\log p(Y_i=k|\cdot)
\end{align}

The left-hand side corresponds to $J_{\text{CE}}$, while the right-hand side is precisely $J_{\text{SNCE}}$.

\subsection{Knowledge Distillation Perspective}

We now discuss the knowledge distillation interpretation.

In standard knowledge distillation, the teacher and student models typically perform the same task (e.g., image classification or object detection). In our setting, however, the teacher model is an implicit distribution $q$ defined by the geometric structure of the continuous latent space.

This paradigm is more analogous to the self-distillation framework explored in recent LLM works~\cite{shenfeld2026self}. In that setup, the model is given a question and demonstration reasoning traces. Instead of directly training with a cross-entropy objective on the demonstrations, the model is prompted to generate its own response:

\begin{verbatim}
<Question>
This is an example response to the question:
<Demonstration>
Now answer with a response of your own.
\end{verbatim}

The generated response and its token probabilities are then used as soft supervision signals.

In discrete image generation, we cannot directly prompt the model to generate images based on demonstration images. However, we can interpret the neighborhood distribution $q$ as an analogue of this process. Sampling from the neighborhood distribution (Figure~\ref{fig:visual_distance}) produces visually similar images represented by different token combinations. Therefore, it is reasonable to interpret $q$ as a form of teacher output.

\subsection{On-Policy Learning Perspective}

We also provide an on-policy learning interpretation of the SNCE loss.

Consider the objective

\begin{align}
J_\pi = \mathbb{E}_{a \sim \pi(a|s)} [r(s,a)]
\end{align}

where $r(s,a)$ is a reward function and $\pi_\theta(a|s)$ is the policy.

The corresponding policy gradient is

\begin{align}
\nabla_\theta J_\pi =
\mathbb{E}_{a \sim \pi(a|s)}
\left[
r(s,a)\nabla_\theta \log \pi_\theta(a|s)
\right]
\end{align}

as shown in~\cite{wu2025diversity}.

If we treat token prediction as a decision-making process with $\pi(a|s)=p_\theta(Y_i=a|\cdot)$ and define the reward

\[
r_{\text{SNCE}}(s,a)=
\frac{q_a(z_i)}{p_\theta(Y_i=a|\cdot)},
\]

then

\begin{align}
\nabla_\theta J_\pi
&=
\mathbb{E}_{a \sim \pi(a|s)}
\left[
\frac{q_a(z_i)}{p_\theta(Y_i=a|\cdot)}
\nabla_\theta \log p_\theta(Y_i=a|\cdot)
\right]
\nonumber\\
&=
\sum_{a=1}^K
q_a(z_i)
\nabla_\theta \log p_\theta(Y_i=a|\cdot)
\nonumber\\
&=
\nabla_\theta
\left(
\sum_{a=1}^K
q_a(z_i)
\log p_\theta(Y_i=a|\cdot)
\right)
\nonumber\\
&=
\nabla_\theta J_{\text{SNCE}}
\end{align}

which exactly matches the gradient of $J_{\text{SNCE}}$.

This derivation closely follows~\cite{wu2025diversity}, which shows that standard cross-entropy with one-hot targets corresponds to on-policy learning with the reward

\[
r_{\text{CE}}(s,a)=
\frac{\mathbf{I}\{a=y_i\}}{p_\theta(Y_i=a|\cdot)}.
\]

\section{Additional Experiment Details and Results}

\subsection{Experiment Setup}

\textbf{Visualization in Figure~\ref{fig:visual_distance}.}
To produce Figure~\ref{fig:visual_distance}, we use one-hot ground-truth targets. For each example, we construct a soft categorical distribution such that

\[
\frac{P(\text{top token})}{P(\text{any other token})}=100.
\]

We then compute the cross-entropy between this distribution and the target. For instance, in the third image from the left (the second-closest token), we set the probability of the second-closest token to be $100\times$ larger than the probability of any other token. This design improves readability because the cross-entropy between two one-hot distributions is $+\infty$ when they differ and $0$ when they match.

\textbf{Toy Example.}
In the toy example, we consider a mixture of two Gaussians centered at $(-2,0)$ and $(2,0)$ with variance $0.25$. The quantization process is defined using a $50\times50$ grid over the square region $[-5,5]\times[-5,5]$. Each point is mapped to its nearest grid point, creating a vocabulary of $2{,}500$ tokens.

We sample 100 data points for training and train small MLPs for 2,000 steps, which we find sufficient for model convergence.

\textbf{ImageNet.}
We follow the model architecture and hyperparameters of IBQ-B~\cite{shi2025scalable}, except that we resize the input embedding and final linear layer to match the enlarged codebook size.

\begin{table}[t]
\centering
\caption{\textbf{Training configurations.} We report the relevant hyperparameters for training, including the learning rate, number of training steps, optimizer setup, image resolution fo generation tasks. }
\label{tab:training-stages}
\begin{tabular}{lHHc}
\toprule

 & \textbf{Stage 1} & \textbf{Stage 2} & \textbf{SFT} \\
\midrule
Learning Rate & $5 \times 10^{-6}$ & $1 \times 10^{-4}$ & $2 \times 10^{-5}$ \\
Steps & 80k & 400k & 100k \\
$\beta_1$ & 0.99 &0.99  & 0.99 \\
$\beta_2$ & 0.999 &0.999  & 0.999 \\
optimizer & AdamW & AdamW & AdamW \\
\midrule
Loaded Parameters & 8B & 6.4B & 10B \\
Trainable Parameters & 8B & 2.4B & 2B \\
Gen. resolution & - & 256 $\rightarrow$ 512 $\rightarrow$ 1024 & 1024 \\
\midrule
\end{tabular}%

\end{table}

\textbf{Large-Scale T2I.}
Our training dataset consists of:

\begin{itemize}
\item \textit{Text-to-image data}: LAION-2B~\cite{schuhmann2022laion}, COYO-700M~\cite{kakaobrain2022coyo-700m}, BLIP3o-60k~\cite{chen2025blip3}, and ShareGPT4o-Image~\cite{chen2025sharegpt}. These datasets are heavily filtered to remove NSFW prompts, low CLIP scores~\cite{radford2021learning}, low aesthetic scores~\cite{laion-aesthetics}, and low-resolution images following the LaViDa-O pipeline.

\item \textit{Image editing data}: ShareGPT4o-Image~\cite{chen2025sharegpt}, GPT-Edit-1.5M~\cite{wang2025gpt}, and UniWorld-V1~\cite{hu2022unified}.
\end{itemize}

In total, we use 50M text-to-image samples and 2M image-editing samples. The training schedule is 200k steps with a global batch size of 1,024. Image-editing data is introduced only during the final 100k steps.

The base model LaViDa-O is a unified multimodal model with separate branches for visual understanding and visual generation. We fine-tune only the visual generation branch while keeping the understanding branch frozen.

The full hyperparameters are listed in Table~\ref{tab:training-stages}.

\subsection{Comparison with Label Smoothing}

Another approach for softening the one-hot target distribution is label smoothing (LS), which defines

\begin{align}
q_k^{\text{LS}}(y_i) =
\begin{cases}
1-\epsilon, & k = y_i \\
\frac{\epsilon}{K-1}, & k \neq y_i .
\end{cases}
\end{align}

However, this approach provides only limited benefits because it assigns a small uniform probability to all non-target tokens without considering the semantic structure of the embedding space. Consequently, semantically adjacent tokens still receive negative gradients (see Equation~\ref{eq:gradient}), since $\frac{\epsilon}{K-1}$ is extremely small.

In contrast, we expect a well-trained generative model to assign relatively high probability to tokens that are close in embedding space, which means they will receive negative gradients even with label smoothing. We emprically compare SNCE and LS on ImageNet 256 and report experiment results in Table~\ref{tab:ls}.

\begin{table}[t]
\centering
\caption{\textbf{Additional Experiment Results on ImageNet256.}}

\begin{subtable}{0.48\linewidth}
\centering
\caption{Effect of temperature $\tau$.}
\label{tab:ablation}
\begin{tabular}{ccc}
\toprule
\textbf{$\tau$} & \textbf{$2\tau^2$} & \textbf{FID} $\downarrow$ \\
\midrule
0.50 & 0.50 & 5.17 \\
0.71 & 1.00 & 3.42 \\
1.00 & 2.00 & 3.46 \\
1.41 & 4.00 & 5.36 \\
\bottomrule
\end{tabular}
\end{subtable}
\hfill
\begin{subtable}{0.48\linewidth}
\centering
\caption{Comparison with label smoothing.}
\label{tab:ls}
\begin{tabular}{cc}
\toprule
\textbf{Method} & \textbf{FID} $\downarrow$ \\
\midrule
CE & 5.44 \\
\hline
CE+LS ($\epsilon=0.05$) & 5.51 \\
CE+LS ($\epsilon=0.1$) & 5.74 \\
\hline
SNCE & 3.42 \\
\bottomrule
\end{tabular}
\end{subtable}

\end{table}

\subsection{Ablation Studies on Temperature $\tau$}
\begin{figure}
\centering
\includegraphics[width=0.9\linewidth]{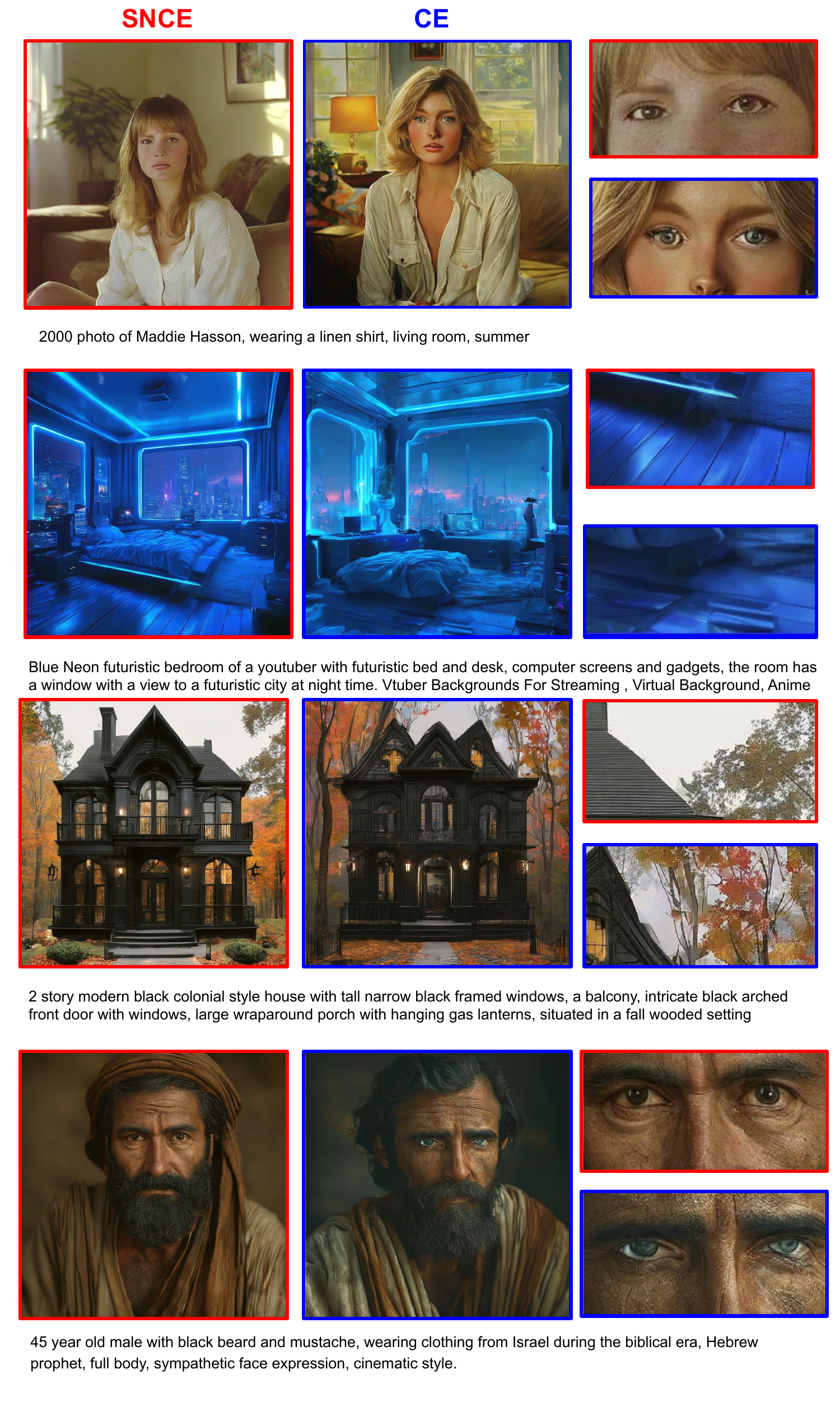}
\caption{\textbf{Additional qualitative comparisons.}}
\label{fig:appendix_t2i}
\end{figure}

An important hyperparameter in SNCE is the temperature $\tau$.

When $\tau$ is small, the neighbor distribution $q$ becomes highly concentrated. As $\tau \rightarrow 0$, $q$ approaches a one-hot distribution and SNCE reduces to standard cross entropy.

When $\tau$ is large, the distribution becomes overly diffuse. As $\tau \rightarrow \infty$, $q$ approaches a uniform distribution, which may lead to a low signal-to-noise ratio in the training signal.

We evaluate several values of $\tau$ on ImageNet and report the results in Table~\ref{tab:ablation}. We find that $\tau=0.71$ achieves the best performance, while both larger and smaller values lead to degraded results.

\subsection{Additional Qualitative Results}

We present additional qualitative results in Figure~\ref{fig:appendix_t2i}. We observe that models trained with the SNCE objective produce higher-quality images than those trained with standard cross entropy, particularly in fine details such as facial features and eyes.

\section{Additional Discussions}

\textbf{Factorized VQ-VAE.}
Some works, such as LFQ~\cite{yu2023language}, recognize the difficulty of training generative models with very large codebooks and address this issue through specially designed tokenizers. These methods quantize image latents using multiple smaller codebooks in a factorized manner rather than a single large codebook.

While such approaches reduce the effective vocabulary size seen by the generative model, they introduce additional structural assumptions on the latent representation and tokenizer architecture.

In contrast, our work focuses on a generic solution for training discrete image generators with a large flat codebook without assuming any factorization structure. This setting is particularly relevant because flat codebooks have been shown to scale well in practice, as demonstrated by recent systems such as Emu3.5~\cite{cui2025emu3}.

\section{Limitations}

Despite the promising results, our work has several limitations. First, while we show that SNCE improves visual quality and text alignment, generated images are still not pixel-perfect and may contain small artifacts.

Second, our model inherits many limitations from the base model LaViDa-O, such as hallucination and soical biases. The trained model is intended for research purpose only and we caution against any other use.

\end{document}